\documentclass{article}
\pdfoutput=1

\usepackage[utf8]{inputenc}
\usepackage{times}
\usepackage{graphicx}
\usepackage{subfigure} \usepackage{natbib}
\usepackage{algorithmic}
\usepackage{hyperref} \usepackage{amsmath}
\usepackage{amssymb}

\newcommand{\DSGF}{\begin{sc}\mbox{DSG-F}\end{sc}}
\newcommand{\DSGS}{\begin{sc}\mbox{DSG-S}\end{sc}}
\newcommand{\HAM}{\begin{sc}SGI\end{sc}}
\newcommand{\KIM}{\begin{sc}SGP\end{sc}}
\usepackage[accepted]{icml2017}

\definecolor{royalblue}{rgb}{0.0, 0.14, 0.4}
\hypersetup{citecolor=royalblue}

\icmltitlerunning{Dynamic Word Embeddings}

\begin{document}

\twocolumn[ \icmltitle{Dynamic Word Embeddings}

\begin{icmlauthorlist}
\icmlauthor{Robert Bamler}{dr}
\icmlauthor{Stephan Mandt}{dr}
\end{icmlauthorlist}

\icmlaffiliation{dr}{Disney Research, 4720 Forbes Avenue, Pittsburgh, PA 15213,
USA}

\icmlcorrespondingauthor{Robert Bamler}{Robert.Bamler@disneyresearch.com}
\icmlcorrespondingauthor{Stephan Mandt}{Stephan.Mandt@disneyresearch.com}

\icmlkeywords{machine learning, natural language processing}

\vskip 0.3in ]

\printAffiliationsAndNotice{}

\begin{abstract}
We present a probabilistic language model for time-stamped text data which tracks the semantic evolution of individual words over time.
The model represents words and contexts by latent trajectories in an embedding space.
At each moment in time, the embedding vectors are inferred from a probabilistic version of word2vec~\citep{mikolov_distributed_2013}. These embedding vectors are connected in time through a latent  diffusion process.
We describe two scalable variational inference algorithms---skip-gram smoothing and skip-gram filtering---that allow us to train the model jointly over all times; thus learning on all data while simultaneously allowing word and context vectors to drift.
Experimental results on three different corpora demonstrate that our dynamic model infers word embedding trajectories that are more interpretable and lead to higher predictive likelihoods than competing methods that are based on static models trained separately on time slices.
\end{abstract}

\section{Introduction}
\label{sec:motivation}

Language evolves over time and words change their meaning due to cultural shifts, technological inventions, or political events. We consider the problem of detecting shifts in the meaning and usage of words over a given time span based on text data.
Capturing these semantic shifts requires a dynamic language model.

Word embeddings are a powerful tool for modeling semantic relations between individual words~\citep{bengio2003neural,mikolov_efficient_2013,pennington2014glove,mnih2013learning,levy2014neural,vilnis2014word,rudolph2016exponential}.
Word embeddings model the distribution of words based on their
surrounding words in a training corpus, and summarize these statistics in terms of
low-dimensional vector representations. Geometric distances between word vectors reflect semantic similarity~\citep{mikolov_efficient_2013} and difference vectors encode semantic and syntactic relations~\citep{mikolov2013linguistic}, which shows that they are sensible representations of language.
Pre-trained word embeddings are useful for various supervised tasks, including sentiment analysis~\citep{socher2013recursive}, semantic parsing~\citep{socher2013parsing}, and computer vision~\citep{fu2016semi}.
As unsupervised models, they have also been used for the exploration of word analogies and linguistics~\citep{mikolov2013linguistic}.

Word embeddings are currently formulated as static models, which assumes that the meaning of any given word is the same across the entire text corpus.
In this paper, we propose a generalization of word embeddings to sequential data, such as corpora of historic texts or streams of text in social media.

Current approaches to learning word embeddings in a dynamic context rely on grouping the data into time bins and training the embeddings separately on these bins~\citep{kim2014temporal,kulkarni2015statistically,hamilton2016diachronic}.
This approach, however, raises three fundamental problems.
First, since word embedding models are non-convex, training them twice on the same data will lead to different results. Thus, embedding vectors at successive times can only be approximately related to each other, and only if the embedding dimension is large~\citep{hamilton2016diachronic}.
Second, dividing a corpus into separate time bins may lead to training sets that are too small to train a word embedding model. Hence, one runs the risk of overfitting to few data whenever the required temporal resolution is fine-grained, as we show in the experimental section.
Third, due to the finite corpus size the learned word embedding vectors are subject to random noise. It is difficult to disambiguate this noise from systematic semantic drifts between subsequent times, in particular over short time spans, where we expect only minor semantic drift.

\begin{figure*}[t!]
\begin{center}
\centerline{\includegraphics[width=\textwidth]{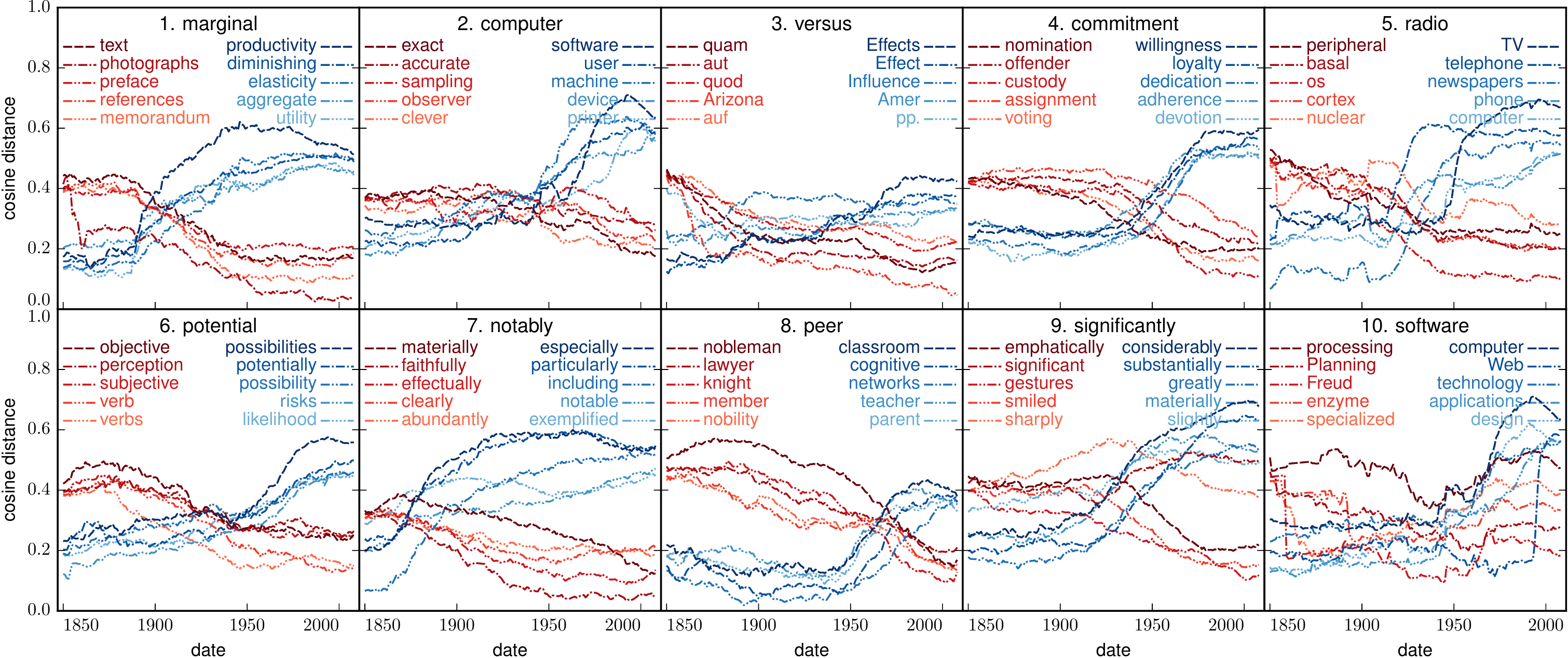}}
\caption{Evolution of the $10$ words that changed the most in cosine distance from 1850 to 2008 on Google books, using skip-gram filtering (proposed).
Red (blue) curves correspond to the five closest words at the beginning (end) of the time span, respectively.
}
\label{fig:timeline2}
\end{center}
\end{figure*}
In this paper, we circumvent these problems by introducing a dynamic word embedding model.
Our contributions are as follows:
\begin{itemize}
\itemsep0em
\item We derive a probabilistic state space model where word and context embeddings evolve in time according to a diffusion process. It generalizes the skip-gram model~\citep{mikolov_distributed_2013,barkan2016bayesian} to a dynamic setup, which allows end-to-end training. This leads to continuous embedding trajectories, smoothes out noise in the word-context statistics, and allows us to share information across all times.
\item We propose two scalable black-box variational inference algorithms~\citep{ranganath2014black,rezende2014stochastic} for filtering and smoothing. These algorithms find word embeddings that generalize better to held-out data. Our smoothing algorithm carries out efficient black-box variational inference for structured Gaussian variational distributions with tridiagonal precision matrices, and applies more broadly.
\item We analyze three massive text corpora that span over long periods of time. Our approach allows us to automatically find the words whose meaning changes the most. It results in smooth word embedding trajectories and therefore allows us to measure and visualize the continuous dynamics of the entire embedding cloud as it deforms over time.
\end{itemize}

Figure~\ref{fig:timeline2} exemplifies our method. The plot shows a fit of our dynamic skip-gram model to Google books (we give details in section~\ref{sec:experiments}). We show the ten words whose meaning changed most drastically in terms of cosine distance over the last $150$ years. We thereby automatically discover words such as ``computer'' or ``radio'' whose meaning changed due to technological advances, but also words like ``peer'' and ``notably'' whose semantic shift is less obvious.

Our paper is structured as follows.
In section \ref{sec:related} we discuss related work, and we introduce our model in section \ref{sec:model}. In section \ref{sec:inference} we present two efficient variational inference algorithms for our dynamic model. We show experimental results in section \ref{sec:experiments}.
Section \ref{sec:conclusion} summarizes our findings.

\section{Related Work}
\label{sec:related}

Probabilistic models that have been extended to latent time series models are ubiquitous~\citep{blei2006dynamic,wang2012continuous,sahoo2012hidden,gultekin_collaborative_2015,charlin2015dynamic,ranganath2015survival,jerfel2016dynamic}, but none of them relate to word embeddings. The closest of these models is
the dynamic topic model~\cite{blei2006dynamic,wang2012continuous}, which learns the evolution of latent topics over time. Topic models are based on bag-of-word representations and thus treat words as symbols without modelling their semantic relations.
They therefore serve a different purpose.

\citet{mikolov_efficient_2013,mikolov_distributed_2013} proposed the skip-gram model with negative sampling (word2vec) as a scalable word embedding approach that relies on stochastic gradient descent. This approach has been formulated in a Bayesian setup~\citep{barkan2016bayesian}, which  we discuss separately in section~\ref{sec:bayesianskip-gram}. These models, however, do not allow the word embedding vectors to change over time.

Several authors have analyzed different statistics of text data to analyze semantic changes of words over time~\citep{mihalcea2012word,sagi2011tracing,kim2014temporal,kulkarni2015statistically,hamilton2016diachronic}. None of them explicitly model a dynamical process; instead, they slice the data into different time bins, fit the model separately on each bin, and further analyze the embedding vectors in post-processing. By construction, these static models can therefore not share statistical strength across time.
This limits the applicability of static models to very large corpora.

Most related to our approach are methods based on word embeddings. \citet{kim2014temporal} fit word2vec separately on different time bins, where the word vectors obtained for the previous bin are used to initialize the algorithm for the next time bin. The bins have to be sufficiently large and the found trajectories are not as smooth as ours, as we demonstrate in this paper.
\citet{hamilton2016diachronic} also trained word2vec separately on several large corpora from different decades. If the embedding dimension is large enough (and hence the optimization problem less non-convex), the authors argue that word embeddings at nearby times approximately differ by a global rotation in addition to a small semantic drift, and they approximately compute this rotation. As the latter does not exist in a strict sense, it is difficult to distinguish artifacts of the approximate rotation from a true semantic drift. As discussed in this paper, both variants result in trajectories which are noisier.\footnote{
    \citet{rudolph2017dynamic} independently developed a similar model, using a different likelihood model. Their approach uses a non-Bayesian treatment of the latent embedding trajectories, which makes the approach less robust to noise when the data per time step is small.
    }

\begin{figure}[t!]
\begin{center}
\centerline{\includegraphics[width=\columnwidth]{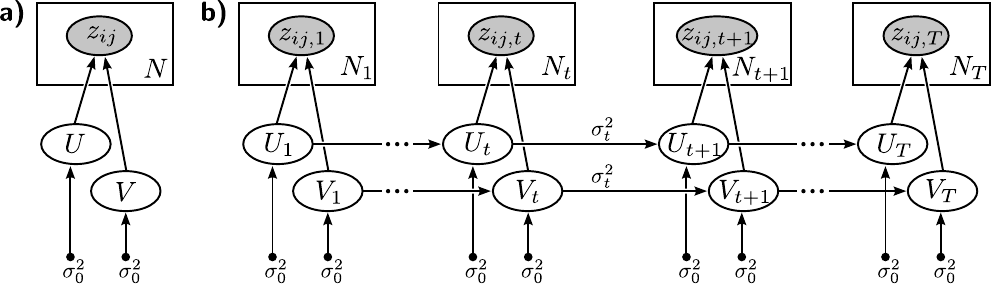}}
\caption{a) Bayesian skip-gram model~\citep{barkan2016bayesian}.
b) The dynamic skip-gram model (proposed) connects $T$ copies of the Bayesian skip-gram model via a latent time series prior on the embeddings.}
\label{fig:graphical-model}
\end{center}
\vskip -0.2in
\end{figure}

\section{Model}
\label{sec:model}
We propose the dynamic skip-gram model, a generalization of the skip-gram model (word2vec) \citep{mikolov_distributed_2013} to sequential text data. The model finds word embedding vectors that continuously drift over time, allowing to track changes in language and word usage over short and long periods of time. Dynamic skip-gram is a probabilistic model which combines a Bayesian version of the skip-gram model~\citep{barkan2016bayesian} with a latent time series. It is jointly trained end-to-end and scales to massive data by means of approximate Bayesian inference.

The observed data consist of sequences of words from a finite vocabulary of size $L$. In section~\ref{sec:bayesianskip-gram}, all sequences (sentences from books, articles, or tweets) are considered time-independent; in section \ref{sec:dynamicskip-gram} they will be associated with different time stamps. The goal is to maximize the probability of every word that occurs in the data given its surrounding words within a so-called context window.
As detailed below, the model learns two vectors $u_i, v_i \in \mathbb R^d$ for each word $i$ in the vocabulary, where $d$ is the embedding dimension.
We refer to $u_i$ as the word embedding vector and to $v_i$ as the context embedding vector.

\subsection{Bayesian Skip-Gram Model}
\label{sec:bayesianskip-gram}

The Bayesian skip-gram model~\citep{barkan2016bayesian} is a probabilistic version of
word2vec~\citep{mikolov_distributed_2013} and forms the basis of our approach.
The graphical model is shown in Figure~\ref{fig:graphical-model}a).
For each pair of words $i,j$ in the vocabulary, the model assigns probabilities that word $i$ appears in the context of word $j$.
This probability is $\sigma(u_i^\top v_j)$ with the sigmoid function $\sigma(x) = 1/(1+e^{-x})$.
Let $z_{ij}\in \{0,1\}$ be an indicator variable that denotes a draw from that probability distribution, hence $p(z_{ij}=1) = \sigma(u_i^\top v_j)$. The generative model assumes that many word-word pairs $(i,j)$ are uniformly drawn from the vocabulary and tested for being a word-context pair; hence a separate random indicator $z_{ij}$ is associated with each drawn pair.

Focusing on words and their neighbors in a context window, we collect evidence of word-word pairs for which $z_{ij}=1$.
These are called the positive examples.
Denote $n^+_{ij}$ the number of times that a word-context pair $(i,j)$ is observed in the corpus.
This is a sufficient statistic of the model, and its contribution to the likelihood is $p(n^+_{ij}|u_i,v_j) = \sigma(u_i^\top v_j)^{n^+_{ij}}$.
However, the generative process also assumes the possibility to reject word-word pairs if $z_{ij}=0$.
Thus, one needs to construct a fictitious second training set of rejected word-word pairs, called negative examples.
Let the corresponding counts be $n^-_{ij}$.
The total likelihood of both positive and negative examples is then
\begin{align}
\label{eq:likelihood}
p(n^+,n^-|U,V) = \prod_{i,j=1}^L \sigma(u_i^\top v_j)^{n^+_{ij}} \sigma(-u_i^\top v_j)^{n^-_{ij}}.
\end{align}
Above we used the antisymmetry $\sigma(-x) = 1-\sigma(x)$.
In our notation, dropping the subscript indices for $n^+$ and $n^-$ denotes the entire $L\times L$ matrices, $U = (u_{1}, \cdots, u_{L}) \in {\mathbb R}^{d\times L}$ is the matrix of all word embedding vectors, and $V$ is defined analogously for the context vectors.
To construct negative examples, one typically chooses $n^-_{ij}\propto P(i)P(j)^{3/4}$~\citep{mikolov_distributed_2013}, where $P(i)$ is the frequency of word $i$ in the training corpus.
Thus, $n^-$ is well-defined up to a constant factor which has to be tuned.

Defining $n^\pm = (n^+,n^-)$ the combination of both positive and negative examples, the resulting log likelihood is
\begin{multline}\label{eq:full-static-likelihood}
\log p(n^\pm|U,V) = \\
\sum_{i,j=1}^{L} \big(n^{+}_{ij} \log \sigma(u_i^\top v_j) +   n^{-}_{ij}\log \sigma(- u_i^\top v_j) \big).
\end{multline}
This is exactly the objective of the (non-Bayesian) skip-gram model, see~\citep{mikolov_distributed_2013}. The count matrices $n^{+}$ and $n^{-}$ are either pre-computed for the entire corpus, or estimated based on stochastic subsamples from the data in a sequential way, as done by word2vec.~\citet{barkan2016bayesian} gives an approximate Bayesian treatment of the model with Gaussian priors on the embeddings.

\subsection{Dynamic Skip-Gram Model}
\label{sec:dynamicskip-gram}

The key extension of our approach is to use a Kalman filter as a prior for the time-evolution of the latent embeddings~\citep{welch1995introduction}. This allows us to share information across all times while still allowing the embeddings to drift.

\paragraph{Notation.}
We consider a corpus of $T$ documents which were written at time stamps $\tau_1 < \ldots < \tau_T$.
For each time step $t\in\{1,\ldots,T\}$ the sufficient statistics of word-context pairs are encoded in the $L\times L$ matrices $n^+_t, n^-_t$ of positive and negative counts with matrix elements $n^+_{ij,t}$ and $n^-_{ij,t}$, respectively.
Denote $U_t = (u_{1,t}, \cdots, u_{L,t}) \in {\mathbb R}^{d\times L}$ the matrix of word embeddings at time $t$, and define $V_t$ correspondingly for the context vectors.
Let $U,V \in {\mathbb R}^{T\times d\times L}$ denote the tensors of word and context embeddings across all times, respectively.

\paragraph{Model.}
The graphical model is shown in Figure~\ref{fig:graphical-model}b).
We consider a diffusion process of the embedding vectors over time.
The variance $\sigma_t^2$ of the transition kernel is
\begin{align}\label{eq:kalman-prior-variance}
\sigma_t^2 = D (\tau_{t+1} - \tau_t),
\end{align}
where $D$ is a global diffusion constant and $(\tau_{t+1}-\tau_t)$ is the time between subsequent observations~\citep{welch1995introduction}. At every time step $t$, we add an additional Gaussian prior with zero mean and variance $\sigma_0^2$ which prevents the embedding vectors from growing very large, thus
\begin{align}
   p(U_{t+1}|U_{t}) \propto {\cal N}(U_{t},\sigma_t^2)\, {\cal N}(0,\sigma_0^2).
\end{align}
Computing the normalization, this results in
\begin{align}
U_{t+1}|U_t \sim {\cal N}\left(\frac{U_t}{1+\sigma_t^2/\sigma_0^2},\frac{1}{\sigma_t^{-2} + \sigma_0^{-2}}I\right), \label{eq:kalman-prior-u}\\
V_{t+1}|V_t \sim {\cal N}\left(\frac{V_t}{1+\sigma_t^2/\sigma_0^2},\frac{1}{\sigma_t^{-2} + \sigma_0^{-2}}I\right).
\label{eq:kalman-prior-v}
\end{align}
In practice, $\sigma_0\gg \sigma_t$, so the damping to the origin is very weak. This is also called Ornstein-Uhlenbeck process~\citep{uhlenbeck_theory_1930}. We recover the Wiener process for $\sigma_0\rightarrow \infty$, but $\sigma_0<\infty$ prevents the latent time series from diverging to infinity.
At time index $t=1$, we define $p(U_1|U_0) \equiv {\cal N}(0,\sigma_0^2I)$ and do the same for $V_1$.

Our joint distribution factorizes as follows:
\begin{multline}
\label{eq:joint}
p(n^\pm,U,V) = \prod_{t=0}^{T-1} p(U_{t+1}|U_t)\,p(V_{t+1}|V_t) \\
\times \prod_{t=1}^T \prod_{i,j=1}^L p(n_{ij,t}^{\pm} | u_{i,t},v_{j,t})
\end{multline}

The prior model enforces that the model learns embedding vectors which vary smoothly across time. This allows to associate words unambiguously with each other and to detect semantic changes. The model efficiently shares information across the time domain, which allows to fit the model even in setups where the data at every given point in time are small, as long as the data in total are large.

\section{Inference}
\label{sec:inference}

We discuss two scalable approximate inference algorithms.
\emph{Filtering} uses only information from the past; it is required in streaming applications where the data are revealed to us sequentially. \emph{Smoothing} is the other inference method, which learns better embeddings but requires the full sequence of documents ahead of time.

In Bayesian inference, we start by formulating a joint distribution (Eq.~\ref{eq:joint}) over observations $n^{\pm}$ and parameters $U$ and $V$, and we are interested in the posterior distribution over parameters conditioned on observations,
\begin{align}
p(U,V | n^\pm) = \frac{p(n^\pm,U,V)}{\int p(n^\pm,U,V)\, dU dV}
\end{align}
The problem is that the normalization is intractable. In variational inference (VI) \citep{jordan_introduction_1999,blei_variational_2016} one sidesteps this problem and approximates the posterior with a simpler variational distribution $q_\lambda(U,V)$ by minimizing the Kullback-Leibler (KL) divergence to the posterior.
Here, $\lambda$ summarizes all parameters of the variational distribution, such as the means and variances of a Gaussian, see below.
Minimizing the KL divergence is equivalent to optimizing the evidence lower bound (ELBO)~\citep{blei_variational_2016},
\begin{multline}\label{eq:elbo}
    \mathcal L(\lambda) = \mathbb E_{q_\lambda}[\log p(n^\pm,U, V)]
    -\mathbb E_{q_\lambda}[\log q_{\lambda}(U,V)].
\end{multline}

For a restricted class of models, the ELBO can be computed in closed-form~\citep{hoffman_stochastic_2013}. Our model is non-conjugate and requires instead black-box VI using the reparameterization trick~\citep{rezende2014stochastic,kingma_autoencoding_2013}.

\subsection{Skip-Gram~Filtering} In many applications such as streaming, the data arrive sequentially. Thus, we can only condition our model on past and not on future observations. We will first describe inference in such a (Kalman) filtering setup~\citep{kalman_new_1960,welch1995introduction}.

In the filtering scenario, the inference algorithm iteratively updates the variational distribution $q$ as evidence from each time step $t$ becomes available. We thereby use a variational distribution that factorizes across all times, $q(U,V) = \prod_{t=1}^T q(U_t,V_t)$ and we update the variational factor at a given time $t$ based on the evidence at time $t$ and the approximate posterior of the previous time step.
Furthermore, at every time $t$ we use a fully-factorized distribution:
\begin{align}
    q(U_t,V_t) =  \prod_{i=1}^L \mathcal N(u_{i,t}; \mu_{ui,t}, \Sigma_{ui,t})\, \mathcal N(v_{i,t}; \mu_{vi,t}. \Sigma_{vi,t}), \nonumber
\end{align}
The variational parameters are the means $\mu_{ui,t},\mu_{vi,t}\in\mathbb R^d$ and the covariance matrices $\Sigma_{ui,t}$ and $\Sigma_{vi,t}$, which we restrict to be diagonal (mean-field approximation).

We now describe how we sequentially compute $q(U_t,V_t)$ and use the result to proceed to the next time step. As other Markovian dynamical systems, our model assumes the following recursion,
\begin{align}
    p(U_t,V_t | n_{1:t}^\pm) \propto p(n_t^\pm | U_t,V_t) \, p(U_t,V_t | n_{1:t-1}^\pm).
\end{align}
Within our variational approximation, the ELBO (Eq.~\ref{eq:elbo}) therefore separates into a sum of $T$ terms, $\mathcal L = \sum_t \mathcal L_t$ with
\begin{multline}
\label{eq:elbo-filtering}
    \mathcal L_t = \mathbb E[\log p(n_t^\pm|U_t, V_t)] + \mathbb E[\log p(U_t, V_t|n^\pm_{1:t-1})] \\
    - \mathbb E[\log q(U_t,V_t)],
\end{multline}
where all expectations are taken under $q(U_t,V_t)$.
We compute the entropy term $-\mathbb E[\log q]$ in Eq.~\ref{eq:elbo-filtering} analytically and estimate the gradient of the log likelihood by sampling from the variational distribution and using the reparameterization trick \citep{kingma_autoencoding_2013,salimans_weight_2016}.
However, the second term of Eq.~\ref{eq:elbo-filtering}, containing the prior at time $t$, is still intractable. We approximate the prior as
\begin{multline}\label{eq:prior-filtering}
    p(U_t,V_t|n^\pm_{1:t-1}) \equiv \\  \mathbb E_{p(U_{t-1},V_{t-1}|n^\pm_{1:t-1})}\big[p(U_t,V_t|U_{t-1},V_{t-1})\big] \\
   \approx \mathbb E_{q(U_{t-1},V_{t-1})}\big[p(U_t,V_t|U_{t-1},V_{t-1})\big].\,
\end{multline}
The remaining expectation involves only Gaussians and can be carried-out analytically. The resulting approximate prior is a fully factorized distribution $p(U_t,V_t|n^\pm_{1:t-1}) \approx \prod_{i=1}^L \mathcal N(u_{i,t}; \tilde\mu_{ui,t}, \tilde\Sigma_{ui,t}) \, \mathcal N(v_{i,t}; \tilde\mu_{vi,t}, \tilde\Sigma_{vit})$ with
\begin{alignat}{1}\label{eq:filtering-prior-mode}
\begin{split}
    \tilde\mu_{ui,t} &= \tilde\Sigma_{ui,t} \left( \Sigma_{ui,t-1} + \sigma_t^2 I \right)^{-1} \mu_{ui,t-1};
    \\
    \tilde\Sigma_{ui,t} &= \left[ \left(\Sigma_{ui,t-1} + \sigma_t^2I\right)^{-1} + (1/\sigma_0^2) I \right]^{-1}.
\end{split}
\end{alignat}
Analogous update equations hold for $\tilde\mu_{vi,t}$ and $\tilde\Sigma_{vi,t}$.
Thus, the second contribution in Eq.~\ref{eq:elbo-filtering} (the prior) yields a closed-form expression. We can therefore compute its gradient.

\subsection{Skip-Gram~Smoothing} In contrast to filtering, where inference is conditioned on past observations until a given time $t$, (Kalman) smoothing performs inference based on the entire sequence of observations $n^\pm_{1{:}T}$. This approach results in smoother trajectories and typically higher likelihoods than with filtering, because evidence is used from both future and past observations.

Besides the new inference scheme, we also use a different variational distribution.
As the model is fitted jointly to all time steps, we are no longer restricted to a variational distribution that factorizes in time.
For simplicity we focus here on the variational distribution for the word embeddings $U$;
the context embeddings $V$ are treated identically.
We use a factorized distribution over both embedding space and vocabulary space,
\begin{align}
    q(U_{1:T}) = \prod_{i=1}^L\prod_{k=1}^d q(u_{ik,1:T}).
\end{align}

In the time domain, our variational approximation is structured.
To simplify the notation we now drop the indices for words $i$ and embedding dimension $k$, hence we write $q(u_{1:T})$ for $q(u_{ik,1:T})$ where we focus on a single factor.
This factor is a multivariate Gaussian distribution in the time domain with tridiagonal precision matrix $\Lambda$,
\begin{align}
    q(u_{1:T}) = {\cal N}(\mu,\Lambda^{-1})
\end{align}
Both the means $\mu = \mu_{1:T}$ and the entries of the tridiagonal precision matrix $\Lambda\in\mathbb R^{T\times T}$ are variational parameters.
This gives our variational distribution the interpretation of a posterior of a Kalman filter~\citep{blei2006dynamic}, which captures correlations in time.

We fit the variational parameters by training the model jointly on all time steps, using black-box VI and the reparameterization trick.
As the computational complexity of an update step scales as $\Theta(L^2)$, we first pretrain the model by drawing minibatches of $L' < L$ random words and $L'$ random contexts from the vocabulary~\citep{hoffman_stochastic_2013}.
We then switch to the full batch to reduce the sampling noise.
Since the variational distribution does not factorize in the time domain we always include all time steps $\{1,\ldots,T\}$ in the minibatch.

We also derive an efficient algorithm that allows us to estimate the reparametrization gradient using $\Theta(T)$ time and memory, while a naive implementation of black-box variational inference with our structured variational distribution would require $\Theta(T^2)$ of both resources.
The main idea is to parametrize $\Lambda = B^\top B$ in terms of its Cholesky decomposition $B$, which is bidiagonal~\citep{kilic_inverse_2013}, and to express gradients of $B^{-1}$ in terms of gradients of $B$.
We use mirror ascent~\citep{ben_ordered_2001,beck_mirror_2003} to enforce positive definiteness of $B$.
The algorithm is detailed in our supplementary material.

\section{Experiments}
\label{sec:experiments}

We evaluate our method on three time-stamped text corpora.
We demonstrate that our algorithms find smoother embedding trajectories than methods based on a static model.
This allows us to track semantic changes of individual words by following nearest-neighbor relations over time.
In our quantitative analysis, we find higher predictive likelihoods on held-out data compared to our baselines.

\paragraph{Algorithms and Baselines.}
We report results from our proposed algorithms from section~\ref{sec:inference} and compare against baselines from section~\ref{sec:related}:
\begin{itemize}
\itemsep0em
\item {\bf \HAM{}} denotes the non-Bayesian skip-gram model with independent random initializations of word vectors~\citep{mikolov_distributed_2013}. We used our own implementation of the model by dropping the Kalman filtering prior and point-estimating embedding vectors. Word vectors at nearby times are made comparable by approximate orthogonal transformations, which corresponds to~\citet{hamilton2016diachronic}.
\item {\bf \KIM{}} denotes the same approach as above, but with  word and context vectors being pre-initialized with the values from the previous year, as in~\citet{kim2014temporal}.
\item {\bf \DSGF{}}: dynamic skip-gram filtering (proposed).
\item {\bf \DSGS{}}: dynamic skip-gram smoothing (proposed).
\end{itemize}

\paragraph{Data and preprocessing.}
Our three corpora exemplify opposite limits both in the covered time span and in the amount of text per time step.
\begin{itemize}
\itemsep0em
\item[1.] We used data from the {\bf Google books} corpus\footnote{\url{http://storage.googleapis.com/books/ngrams/books/datasetsv2.html}}~\citep{michel_quantitative_2011} from the last two centuries ($T=209$). This amounts to $5$ million digitized books and approximately $10^{10}$ observed words. The corpus consists of $n$-gram tables with $n\in\{1,\ldots,5\}$, annotated by year of publication.
We considered the years from $1800$ to $2008$ (the latest available). In $1800$, the size of the data is approximately $\sim 7\cdot 10^7$ words.
We used the $5$-gram counts, resulting in a context window size of $4$.

\item[2.] We used the ``State~of~the~Union'' ({\bf SoU}) addresses of U.S.~presidents, which spans more than two centuries, resulting in $T=230$ different time steps and approximately $10^6$ observed words.\footnote{\url{http://www.presidency.ucsb.edu/sou.php}}
Some presidents gave both a written and an oral address; if these were less than a week apart we concatenated them and used the average date.
We converted all words to lower case and constructed the positive sample counts $n^+_{ij}$ using a context window size of $4$.

\item[3.] We used a {\bf Twitter} corpus of news tweets for $21$ randomly drawn dates from $2010$ to $2016$.
The median number of tweets per day is $722$.
We converted all tweets to lower case and used a context window size of $4$, which we restricted to stay within single tweets.
\end{itemize}

\paragraph{Hyperparameters.}
The vocabulary for each corpus was constructed from the $10{,}000$ most frequent words throughout the given time period.
In the Google books corpus, the number of words per year grows by a factor of $200$ from the year $1800$ to $2008$.
To avoid that the vocabulary is dominated by modern words we normalized the word frequencies separately for each year before adding them up.

For the Google books corpus, we chose the embedding dimension $d=200$, which was also used in~\citet{kim2014temporal}.
We set $d=100$ for SoU and Twitter, as $d=200$ resulted in overfitting on these much smaller corpora.
The ratio $\eta = \sum_{ij}n_{ij,t}^- / \sum_{ij}n_{ij,t}^+$ of negative to positive word-context pairs was $\eta=1$. The precise construction of the matrices $n^\pm_t$ is explained in the supplementary material.
We used the global prior variance $\sigma_0^2=1$ for all corpora and all algorithms, including the baselines.
The diffusion constant $D$ controls the time scale on which information is shared between time steps.
The optimal value for $D$ depends on the application.
A single corpus may exhibit semantic shifts of words on different time scales, and the optimal choice for $D$ depends on the time scale in which one is interested.
We used $D=10^{-3}$ per year for Google books and SoU, and $D=1$ per year for the Twitter corpus, which spans a much shorter time range.
In the supplementary material, we provide details of the optimization procedure.

\paragraph{Qualitative results.}
We show that our approach results in smooth word embedding trajectories on all three corpora. We can automatically detect  words that undergo significant semantic changes over time.

\begin{figure}[t!]
\begin{center}
\centerline{\includegraphics[width=\columnwidth]{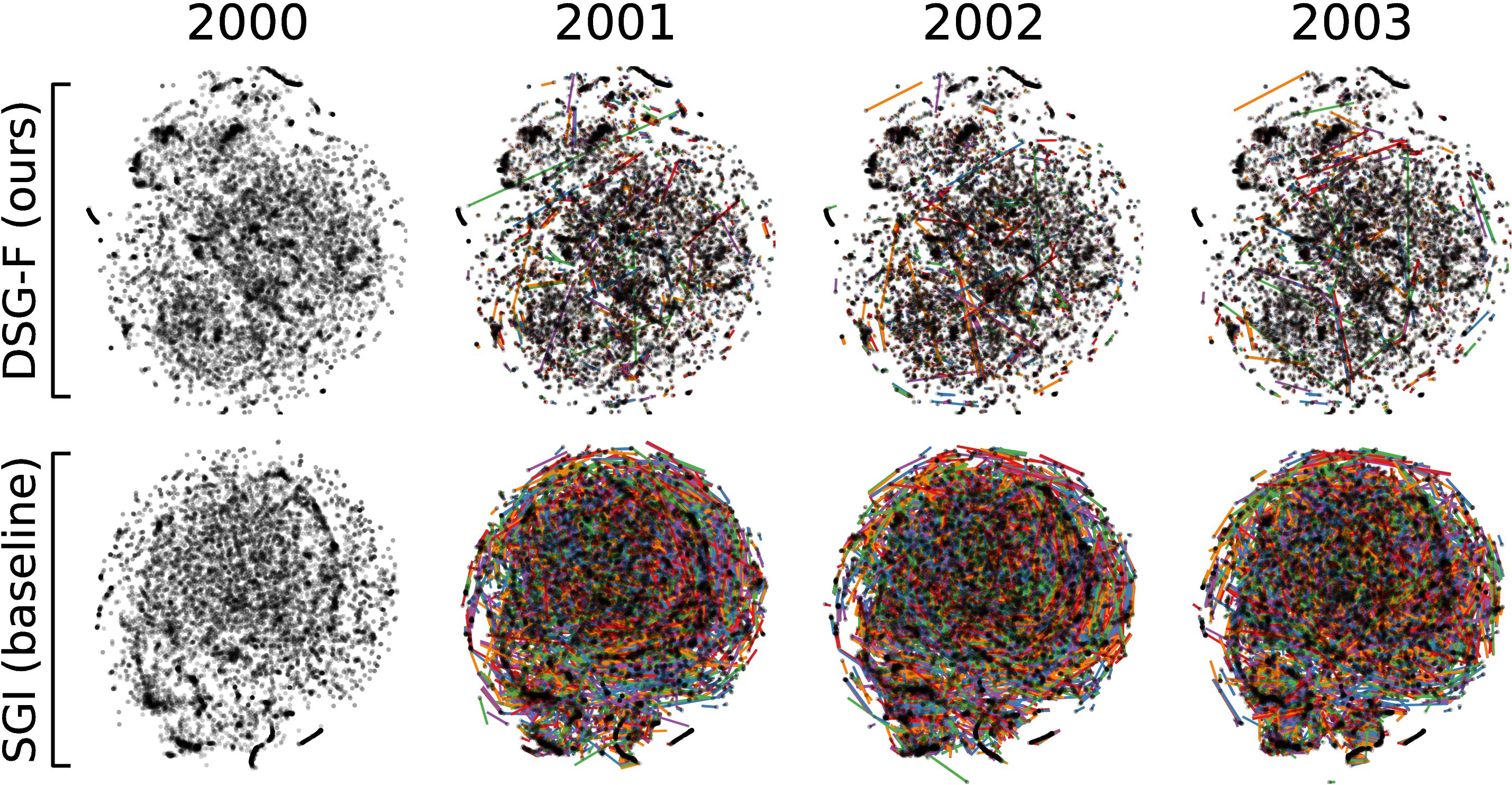}}
\caption{Word embeddings over a sequence of years trained on Google books, using \DSGF{} (proposed, top row) and compared to the static method by~\citet{hamilton2016diachronic} (bottom). We used dynamic t-SNE~\citep{rauber_visualizing_2016} for dimensionality reduction.
Colored lines in the second to fourth column indicate the trajectories from the previous year.
Our method infers smoother trajectories with only few words that move quickly.
Figure~\ref{fig:hist-gbooks} shows that these effects persist in the original embedding space.}
\label{fig:tsne}
\end{center}
\vskip -0.2in
\end{figure}

\begin{figure}[t]
\begin{center}
\centerline{\includegraphics[width=\columnwidth]{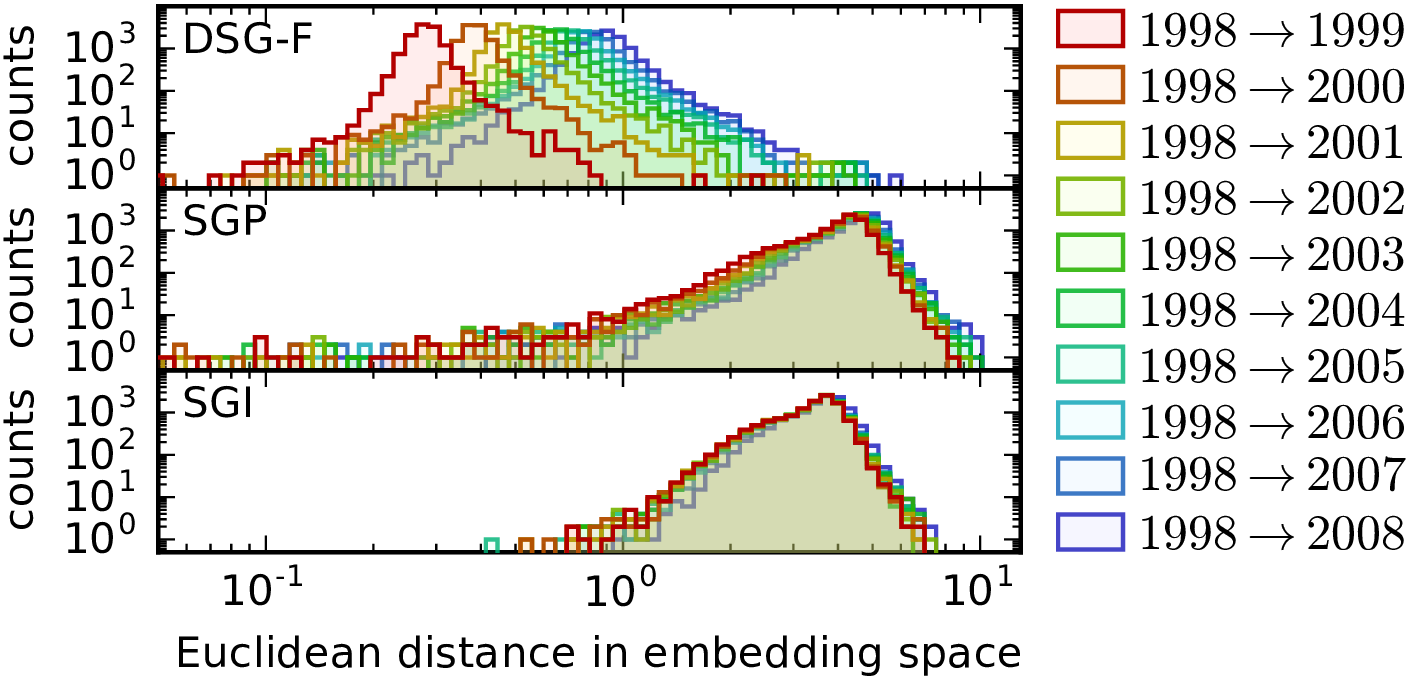}}
\caption{
Histogram of distances between word vectors in the year 1998 and their positions in subsequent years (colors). \DSGF{} (top panel) displays a continuous growth of these distances over time, indicating a directed motion.
In contrast, in \KIM{} (middle)~\citep{kim2014temporal} and \HAM{} (bottom)~\citep{hamilton2016diachronic},
the distribution of distances jumps from the first to the second year but then remains largely stationary, indicating absence of a directed drift; i.e. almost all motion is random.}
\label{fig:hist-gbooks}
\end{center}
\vskip -0.2in
\end{figure}

Figure~\ref{fig:timeline2} in the introduction shows a fit of the dynamic skip-gram filtering algorithm to the Google books corpus. Here, we show the ten words whose word vectors change most drastically over the last $150$ years in terms of cosine distance.
Figure~\ref{fig:tsne} visualizes word embedding clouds over four subsequent years of Google books, where we compare \DSGF{} against \HAM{}.
We mapped the normalized embedding vectors to two dimensions using dynamic t-SNE~\citep{rauber_visualizing_2016} (see supplement for details).
Lines indicate shifts of word vectors relative to the preceding year.
In our model only few words change their position in the embedding space rapidly, while embeddings using \HAM{} show strong fluctuations, making the cloud's motion hard to track.

Figure~\ref{fig:hist-gbooks} visualizes the smoothness of the trajectories directly in the embedding space (without the projection to two dimensions).
We consider differences between word vectors in the year 1998 and the subsequent 10 years. In more detail, we compute histograms of the Euclidean distances $||u_{it} - u_{i,t+\delta}||$ over the word indexes $i$, where $\delta = 1,\dots,10$ (as discussed previously, \HAM{} uses a global rotation to optimally align embeddings first).
In our model, embedding vectors gradually move away from their original position as time progresses, indicating a directed motion.
In contrast, both baseline models show only little directed motion after the first time step,
suggesting that most temporal changes are due to finite-size fluctuations of $n^{\pm}_{ij,t}$.
Initialization schemes alone, thus, seem to have a minor effect on smoothness.

\begin{figure*}[tb!]
\begin{center}
\centerline{\includegraphics[width=\textwidth]{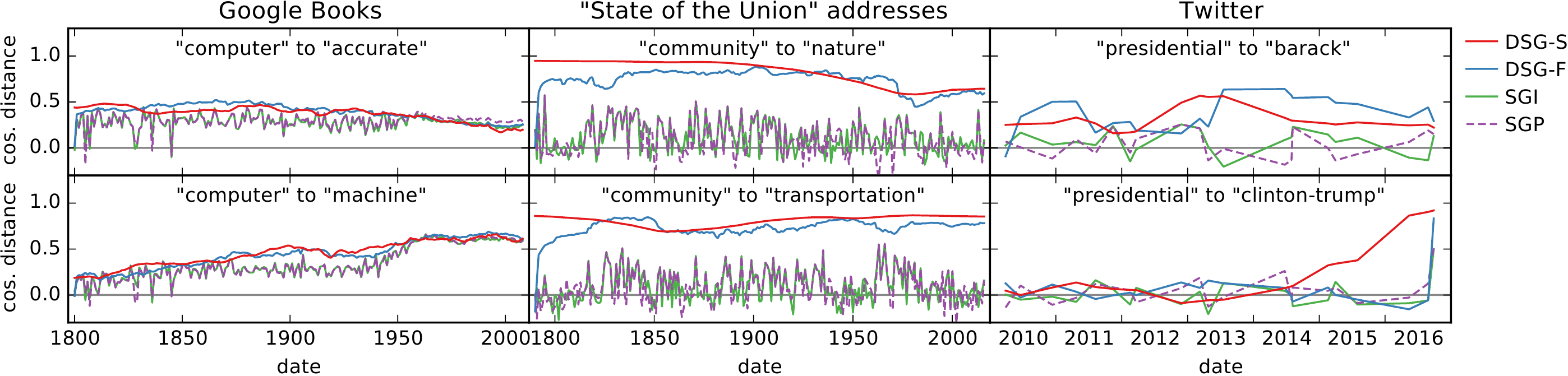}}
\caption{Smoothness of word embedding trajectories, compared across different methods.
We plot the cosine distance between two words (see captions) over time.
High values indicate similarity.
Our methods (\DSGS{} and \DSGF{}) find more interpretable trajectories than the baselines (\HAM{} and \KIM{}).
The different performance is most pronounced when the corpus is small (SoU and Twitter).}
\label{fig:timeline1}
\end{center}
\vskip -0.2in
\end{figure*}

\begin{figure*}[bt!]
\begin{center}
\centerline{\includegraphics[width=\textwidth]{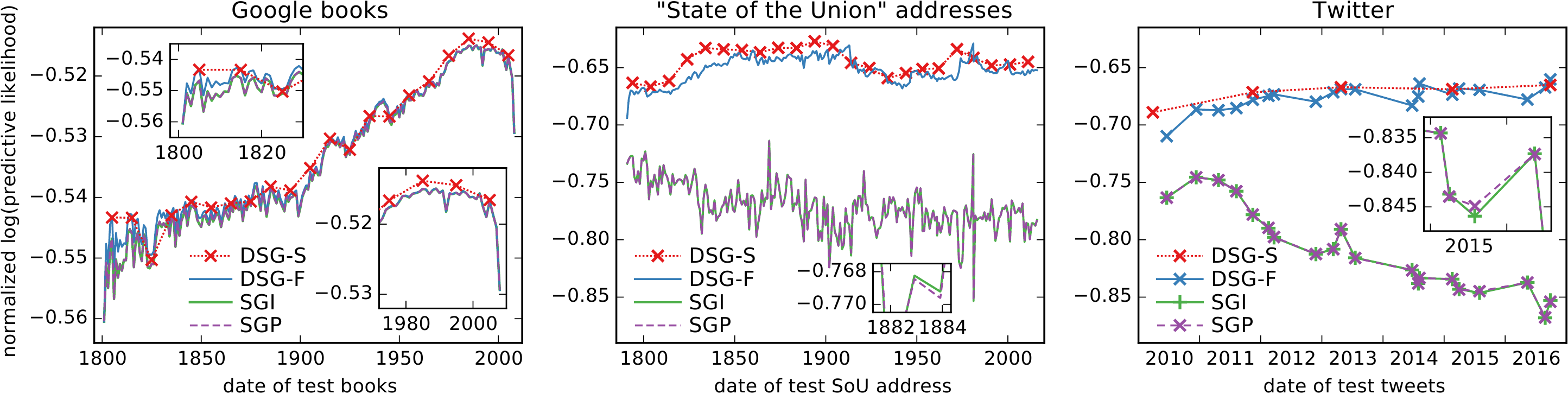}}
\caption{Predictive log-likelihoods (Eq.~\ref{eq:pred-log-likelihood}) for two proposed versions of the dynamic skip-gram model (DSG-F \& DSG-S) and two competing methods \HAM{}~\citep{hamilton2016diachronic} and \KIM{}~\citep{kim2014temporal} on three different corpora (high values are better).
}
\label{fig:pred-log-likelihoods}
\end{center}
\vskip -0.2in
\end{figure*}

Our approach allows us to detect semantic shifts in the usage of specific words.
Figures~\ref{fig:timeline1} and \ref{fig:timeline2} both show the cosine distance between a given
word and its neighboring words (colored lines) as a function of time. Figure~\ref{fig:timeline1} shows results on all
three corpora and focuses on a comparison across methods. We see that \DSGS{} and \DSGF{} (both proposed) result in trajectories which display less noise than the baselines \KIM{} and \HAM{}. The fact that the baselines predict zero cosine distance (no correlation) between the chosen word pairs on the SoU and Twitter corpora suggests that these corpora are too small to successfully fit these models, in contrast to our approach which shares information in the time domain. Note that as in dynamic topic models, skip-gram smoothing (DSG-S) may diffuse information into the past (see "presidential" to "clinton-trump" in Fig.~\ref{fig:timeline1}).

\paragraph{Quantitative results.}
We show that our approach generalizes better to unseen data. We thereby analyze held-out predictive likelihoods on word-context pairs at a given time $t$, where $t$ is excluded from the training set,
\begin{align}\label{eq:pred-log-likelihood}
   \textstyle \frac{1}{|n_t^{\pm}|} \log p(n_t^\pm | \tilde U_t, \tilde V_t).
\end{align}
Above, $|n_t^{\pm}| =\sum_{i,j}\left(n_{ij,t}^+ + n_{ij,t}^- \right)$ denotes the total number of word-context pairs at time $\tau_t$. Since inference is different in all approaches, the definitions of word and context embedding matrices $\tilde U_t$ and $\tilde V_t$ in Eq.~\ref{eq:pred-log-likelihood} have to be adjusted:

\begin{itemize}
\itemsep0em
\item For \HAM{} and \KIM{}, we did a chronological pass through the time sequence and used the embeddings ${\tilde U}_t = U_{t-1}$ and ${\tilde V}_t = V_{t-1}$ from  the previous time step to predict the statistics $n^\pm_{ij,t}$ at time step $t$.
\item For \DSGF{}, we did the same pass to test $n^\pm_{ij,t}$.  We thereby set $\tilde U_t$ and $\tilde V_t$ to be the modes $U_{t-1},V_{t-1}$ of the approximate posterior at the previous time step.
\item For \DSGS{}, we held out $10\%$, $10\%$ and $20\%$ of the documents from the Google books, SoU, and Twitter corpora for testing, respectively.  After training, we estimated the word (context) embeddings $\tilde U_t$ ($\tilde V_t$) in Eq.~\ref{eq:pred-log-likelihood} by linear interpolation between the values of $U_{t-1}$ ($V_{t-1}$) and $U_{t+1}$ ($V_{t+1}$) in the mode of the variational distribution, taking into account that the time stamps $\tau_t$ are in general not equally spaced.
\end{itemize}

The predictive likelihoods as a function of time $\tau_t$ are shown in Figure~\ref{fig:pred-log-likelihoods}.
For the {Google Books} corpus (left panel in figure \ref{fig:pred-log-likelihoods}), the predictive log-likelihood  grows over time with all four methods.
This must be an artifact of the corpus since \HAM{} does not carry any information of the past.
A possible explanation is the growing number of words per year from $1800$ to $2008$ in the Google Books corpus.
On all three corpora,
differences between the two implementations of the static model (\HAM{} and \KIM{}) are small, which suggests that pre-initializing the embeddings with the previous result may improve their continuity but seems to have little impact on the predictive power.
Log-likelihoods for the skip-gram filter (DSG-F) grow over the first few time steps as the filter sees more data, and then saturate.
The improvement of our approach over the static model is particularly pronounced in the SoU and Twitter corpora, which are much smaller than the massive Google books corpus. There, sharing information between across time is crucial because there is little data at every time slice.
Skip-gram smoothing outperforms skip-gram filtering as it shares information in both time directions and uses a more flexible variational distribution.

\section{Conclusions}
\label{sec:conclusion}
We presented the dynamic skip-gram model: a Bayesian probabilistic model
that combines word2vec with a latent continuous time series.
We showed experimentally that both dynamic skip-gram filtering (which conditions only on past observations)
and dynamic skip-gram smoothing (which uses all data) lead to smoothly changing embedding vectors that  are better
at predicting word-context statistics at held-out time steps.
The benefits are most drastic when the data at individual time
steps is small, such that fitting a static word embedding model is hard.
Our approach may be used as a data mining and anomaly detection tool when streaming text on social media, as well as a tool for historians and social scientists interested in the evolution of language.

\subsection*{Acknowledgements}
We would like to thank Marius Kloft, Cheng Zhang, Andreas Lehrmann, Brian McWilliams, Romann Weber, Michael Clements, and Ari Pakman for valuable feedback.

\bibliography{references} \bibliographystyle{icml2017}

\end{document}


\renewcommand{\theequation}{S\arabic{equation}}
\renewcommand{\thepage}{S\arabic{page}}
\renewcommand{\thetable}{S\arabic{table}}
\renewcommand{\thefigure}{S\arabic{figure}}

\twocolumn[ \icmltitle{Supplementary Material to ``Dynamic Word Embeddings''}

\begin{icmlauthorlist}
\icmlauthor{Robert Bamler}{dr}
\icmlauthor{Stephan Mandt}{dr}
\end{icmlauthorlist}

\icmlaffiliation{dr}{Disney Research, 4720 Forbes Avenue, Pittsburgh, PA 15213,
USA}

\icmlcorrespondingauthor{Robert Bamler}{Robert.Bamler@disneyresearch.com}
\icmlcorrespondingauthor{Stephan Mandt}{Stephan.Mandt@disneyresearch.com}

\icmlkeywords{machine learning, natural language processing}

\vskip 0.3in ]

\printAffiliationsAndNotice{}

\begin{table}[tb]
\caption{Hyperparameters for skip-gram filtering and skip-gram smoothing.}
\label{tab:hyperparameters}
\vskip 0.15in
\begin{center}
\begin{small}
\begin{tabular}{@{$\,$}r@{$\,$}c@{$\,$}l@{$\quad$}l@{}}
\hline
\multicolumn{3}{@{}c@{}}{\abovespace\belowspace \begin{sc}Parameter\end{sc}} & \begin{sc}Comment\end{sc}  \\
\hline
\abovespace
$L$ & $=$ & $10^4$ & vocabulary size \\
$L'$ & $=$ & $10^3$ & batch size for smoothing \\
$d$ & $=$ & $100$ & embedding dimension for SoU and Twitter \\
$d$ & $=$ & $200$ & embedding dimension for Google books \\
$N_\text{tr}$ & $=$ & $5000$ & number of training steps for each $t$ (filtering) \\
$N_\text{tr}'$ & $=$ & $5000$ & number of pretraining steps with minibatch \\
& & & sampling (smoothing; see Algorithm~\ref{alg:smoothing}) \\
$N_\text{tr}$ & $=$ & $1000$ & number of training steps without minibatch \\
& & & sampling (smoothing; see Algorithm~\ref{alg:smoothing}) \\
$c_\text{max}$ & $=$ & $4$ & context window size for positive examples \\
$\eta$ & $=$ & $1$ & ratio of negative to positive examples \\
$\gamma$ & $=$ & $0.75$ & context exponent for negative examples \\
$D$ & $=$ & $10^{-3}$ & diffusion const. per year (Google books \& SoU) \\
$D$ & $=$ & $1$ & diffusion const. per year (Twitter) \\
$\sigma_0^2$ & $=$ & $1$ & variance of overall prior \\
$\alpha$ & $=$ & $10^{-2}$ & learning rate (filtering) \\
$\alpha'$ & $=$ & $10^{-2}$ & learning rate during minibatch phase (smoothing) \\
$\alpha$ & $=$ & $10^{-3}$ & learning rate after minibatch phase (smoothing) \\
$\beta_1$ & $=$ & $0.9$ & decay rate of 1\textsuperscript{st} moment estimate \\
$\beta_2$ & $=$ & $0.99$ & decay rate of 2\textsuperscript{nd} moment estimate (filtering)\\
$\beta_2$ & $=$ & $0.999$ & decay rate of 2\textsuperscript{nd} moment estimate (smoothing)\\
\belowspace
$\delta$ & $=$ & $10^{-8}$ & regularizer of Adam optimizer \\
\hline
\end{tabular}
\end{small}
\end{center}
\vskip -0.1in
\end{table}

\section{Dimensionality Reduction in Figure~1}

To create the word-clouds in Figure~1 of the main text we mapped the fitted word embeddings from $\mathbb R^d$ to the two-dimensional plane using dynamic t-SNE~\citep{rauber_visualizing_2016}.
Dynamic t-SNE is a non-parametric dimensionality reduction algorithm for sequential data.
The algorithm finds a projection to a lower dimension by solving a non-convex optimization problem that aims at preserving nearest-neighbor relations at each individual time step.
In addition, projections at neighboring time steps are aligned with each other by a quadratic penalty with prefactor $\lambda\geq 0$ for sudden movements.

There is a trade-off between finding good local projections for each individual time step ($\lambda\to0$), and finding smooth projections (large $\lambda$).
Since we want to analyze the smoothness of word embedding trajectories, we want to avoid bias towards smooth projections.
Unfortunately, setting $\lambda=0$ is not an option since, in this limit, the optimization problem is invariant under independent rotations at each time, rendering trajectories in the two-dimensional projection plane meaningless.
To still avoid bias towards smooth projections, we anneal $\lambda$ exponentially towards zero over the course of the optimization.
We start the optimizer with $\lambda=0.01$, and we reduce $\lambda$ by $5\%$ with each training step.
We run $100$ optimization steps in total, so that $\lambda\approx 6\times 10^{-6}$ at the end of the training procedure.
We used the open-source implementation,\footnote{\url{https://github.com/paulorauber/thesne}} set the target perplexities to $200$, and used default values for all other parameters.

\section{Hyperparemeters and Construction of $n^\pm_{1:T}$}
\label{sec:smoothing}

Table~\ref{tab:hyperparameters} lists the hyperparameters used in our experiments.
For the Google books corpus, we used the same context window size $c_\text{max}$ and embedding dimension $d$ as in \citep{kim2014temporal}.
We reduced $d$ for the SoU and Twitter corpora to avoid overfitting to these much smaller data sets.

In constrast to word2vec, we construct our positive and negative count matrices $n^\pm_{ij,t}$ deterministically in a preprocessing step. As detailed below, this is done such that it resembles as closely as possible the stochastic approach in word2vec~\citep{mikolov_distributed_2013}. In every update step, word2vec stochastically samples a context window size uniformly in an interval $[1,\cdots,c_{max}]$, thus the context size fluctuates and nearby words appear more often in the same context than words that are far apart from each other in the sentence. We follow a deterministic scheme that results in similar statistics.
For each pair of words $(w_1, w_2)$ in a given sentence, we
increase the counts $n^+_{i_{w_1}j_{w_2}}$ by $\max\left(0, 1-k/c_\text{max}\right)$, where $0\leq k \leq c_{max}$ is the number of words that appear between $w_1$ and $w_2$, and $i_{w_1}$ and $j_{w_2}$ are the words' unique  indices in the vocabulary.

We also used a deterministic variant of word2vec to construct the negative count matrices $n^-_t$.
In word2vec, $\eta$ negative samples $(i,j)$ are drawn for each positive sample $(i,j')$ by drawing $\eta$ independent values for $j$ from a distribution $P_t'(j)$ defined below.
We define $n^-_{ij,t}$ such that it matches the expectation value of the number of times that word2vec would sample the negative word-context pair $(i,j)$.
Specifically, we define
\begin{align}
    P_t(i) &= \frac{\sum_{j=1}^L n^+_{ij,t}}{
    \sum_{i',j=1}^L n^+_{i'j,t}}, \label{eq:def_pofi} \\
    P'_t(j) &= \frac{\big(P_t(j)\big)^{\!\gamma}}{\sum_{j'=1}^L \big(P_t(j')\big)^{\!\gamma}}, \\
    n_{ij,t}^- &= \bigg(\sum_{i',j'=1}^L n_{i'j',t}^+\bigg) \eta P_t(i) P'_t(j). \label{eq:def_nminus}
\end{align}
We chose $\gamma=0.75$ as proposed in~\citep{mikolov_distributed_2013}, and we set $\eta=1$.
In practice, it is not necessary to explicitly construct the full matrices $n_{t}^-$, and it is more efficient to keep only the distributions $P_t(i)$ and $P'_t(j)$ in memory.

\begin{algorithm}[tb]
\caption{Skip-gram filtering; see section 4 of the main text.}
\label{alg:filtering}
\begin{algorithmic}
\STATE {\bfseries Remark:} All updates are analogous for word and context vectors; we drop their indices for simplicity.
\STATE {\bfseries Input:} number of time steps $T$, time stamps $\tau_{1:T}$, positive and negative examples $n^\pm_{1:T}$, hyperparameters.
\vskip 3pt
\STATE Init. prior means $\tilde\mu_{ik,1} \leftarrow 0$ and variances $\tilde\Sigma_{i,1} = I_{d\times d}$
\STATE Init. variational means $\mu_{ik,1} \leftarrow 0$ and var. $\Sigma_{i,1} = I_{d\times d}$
\FOR{$t=1$ {\bfseries to} $T$}
\IF{$t\neq1$}
\STATE Update approximate Gaussian prior with params. $\tilde\mu_{ik,t}$ and $\tilde\Sigma_{i,t}$ using  $\mu_{ik,t-1}$ and $\Sigma_{i,t-1}$, see Eq.~13.
\ENDIF
\STATE Compute entropy $ \mathbb E_q[\log q(\cdot)]$ analytically.
\STATE Compute expected log Gaussian prior with parameters $\tilde{\mu}_{ik,t}$ and $\tilde{\Sigma}_{k,t}$ analytically.
\STATE Maximize $\mathcal L_t$ in Eq.~11, using black-box VI with the reparametrization trick.
\STATE Obtain $\mu_{ik,t}$ and $\Sigma_{i,t}$ as outcome of the optimization.
\ENDFOR
\end{algorithmic}
\end{algorithm}

\section{Skip-gram Filtering Algorithm}

The skip-gram filtering algorithm is described in section 4 of the main text.
We provide a formulation in pseudocode in Algorithm~\ref{alg:filtering}.

\begin{algorithm}[tb]
\caption{Skip-gram smoothing; see section \ref{sec:smoothing}. We drop indices $i$, $j$, and $k$ for word, context, end embedding dimension, respectively, when they are clear from context.}
\label{alg:smoothing}
\begin{algorithmic}
\STATE {\bfseries Input:} number of time steps $T$, time stamps $\tau_{1:T}$, word-
\STATE $\quad$context counts $n^+_{1:T}$, hyperparameters in Table~\ref{tab:hyperparameters}
\vskip 2pt
\STATE Obtain $n_{t}^-\;\forall t$ using Eqs.~\ref{eq:def_pofi}--\ref{eq:def_nminus}
\STATE Initialize $\mu_{u,1:T},\mu_{v,1:T} \leftarrow 0$
\STATE Initialize $\nu_{u,1:T}$, $\nu_{v,1:T}$, $\omega_{u,1:T-1}$, and $\omega_{v,1:T-1}$ such
\STATE $\quad$that $B_u^\top B_u = B_v^\top B_v = \Pi$ (see Eqs.~\ref{eq:bmatrix} and \ref{eq:prior-precision})
\vskip 2pt
\FOR{$step=1$ {\bfseries to} $N'_\text{tr}$}
\STATE Draw $\mathcal I \subset \{1,\ldots,L'\}$ with $|\mathcal I|=L'$ uniformly
\STATE Draw $\mathcal J \subset \{1,\ldots,L'\}$ with $|\mathcal J|=L'$ uniformly
\FORALL{$i\in\mathcal I$}
\STATE Draw $\epsilon_{ui,1:T}^{[s]} \sim \mathcal N(0,I)$
\STATE Solve $B_{u,i} x_{ui,1:T} = \epsilon_{ui,1:T}$ for $x_{ui,1:T}$
\ENDFOR
\STATE Obtain $x_{vj,1:T}$ by repeating last loop $\forall j\in\mathcal J$
\STATE Calculate gradient estimates of $\mathcal L$ for minibatch
\STATE $\quad(\mathcal I,\mathcal J)$ using Eqs.~\ref{eq:grad-mu}, \ref{eq:grad-nu}, and \ref{eq:grad-omega}
\STATE Obtain update steps $d[\cdot]$ for all variational parameters
\STATE $\quad$using Adam optimizer with parameters in Table~\ref{tab:hyperparameters}
\STATE Update $\mu_{u,1:T} \leftarrow \mu_{u,1:T} + d[\mu_{u,1:T}]$, and analogously
\STATE $\quad$for $\mu_{v,1:T}$, $\omega_{u,1:T-1}$, and $\omega_{v,1:T-1}$
\STATE Update $\nu_{u,1:T}$ and $\nu_{v,1:T}$ according to Eq.~\ref{eq:mirrorascent}
\ENDFOR
\vskip 2pt
\STATE Repeat above loop for $N_\text{tr}$ more steps, this time without
\STATE $\quad$minibatch sampling (i.e., setting $L'=L$)
\end{algorithmic}
\end{algorithm}

\section{Skip-gram Smoothing Algorithm}
\label{sec:smoothing}

In this section, we give details for the skip-gram smoothing algorithm, see section 4 of the main text.
A summary is provided in Algorithm \ref{alg:smoothing}.

\paragraph{Variational distribution.}
For now, we focus on the word embeddings, and we simplify the notation by dropping the indices for the vocabulary and embedding dimensions.
The variational distribution for a single embedding dimension of a single word embedding trajectory is
\begin{align}\label{eq:variational-q}
    q(u_{1:T})=\mathcal N(\mu_{u,1:T}, (B_u^\top B_u)^{-1}).
\end{align}
Here, $\mu_{u,1:T}$ is the vector of mean values, and $B_u$ is the Cholesky decomposition of the precision matrix.
We restrict the latter to be bidiagonal,
\begin{align}\label{eq:bmatrix}
    B_u = \begin{pmatrix}
        \nu_{u,1} & \omega_{u,1} & & & \\
        & \nu_{u,2} & \omega_{u,2} & & \\
        & & \ddots & \ddots & \\
        & & & \nu_{u,T-1} & \omega_{u,T-1} \\
        & & & & \nu_T
    \end{pmatrix}
\end{align}
with $\nu_{u,t} >0$ for all $t\in\{1,\ldots,T\}$.
The variational parameters are $\mu_{u,1:T}$, $\nu_{u,1:T}$, and $\omega_{1:T-1}$.
The variational distribution of the context embedding trajectories $v_{1:T}$ has the same structure.

With the above form of $B_u$, the variational distribution is a Gaussian with an arbitrary tridiagonal symmetric precision matrix $B_u^\top B_u$.
We chose this variational distribution because it is the exact posterior of a hidden time-series model with a Kalman filtering prior and Gaussian noise~\citep{blei2006dynamic}.
Note that our variational distribution is a generalization of a fully factorized (mean-field) distribution, which is obtained for $\omega_{u,t}=0\;\forall t$.
In the general case, $\omega_{u,t}\neq 0$, the variational distribution can capture correlations between all time steps, with a dense covariance matrix $(B_u^\top B_u)^{-1}$.

\paragraph{Gradient estimation.}
The skip-gram smoothing algorithm uses stochastic gradient ascent to find the variational parameters that maximize the ELBO,
\begin{align}\label{eq:elbo-split}
    \mathcal L = \mathbb E_q\big[\!\log p(U_{1:T}, V_{1:T}, n^\pm_{1:T}) \big] - \mathbb E_q\big[\!\log q(U_{1:T}, V_{1:T}) \big].
\end{align}
Here, the second term is the entropy, which can be evaluated analytically.
We obtain for each component in vocabulary and embedding space,
\begin{align}
     - \mathbb E_q[\log q(u_{1:T})]
    &= -\sum_{t}\log(\nu_{u,t}) + \text{const.} \label{eq:def-entropy}
\end{align}
and analogously for $-E_q[\log q(v_{1:T})]$.

The first term on the right-hand side of Eq.~\ref{eq:elbo-split} cannot be evaluated analytically.
We approximate its gradient by sampling from $q$ using the reparameterization trick~\citep{kingma_autoencoding_2013,rezende2014stochastic}.
A naive calculation would require $\Omega(T^2)$ computing time since the derivatives of $\mathcal L$ with respect to $\nu_{u,t}$ and $\omega_{u,t}$ for each $t$ depend on the count matrices $n^\pm_{t'}$ of all $t'$.
However, as we show next, there is a more efficient way to obtain all gradient estimates in $\Theta(T)$ time.

We focus again on a single dimension of a single word embedding trajectory $u_{1:T}$, and we drop the indices $i$ and $k$.
We draw $S$ independent samples $u_{1:T}^{[s]}$ with $s\in\{1,\ldots,S\}$ from $q(u_{1:T})$ by parameterizing
\begin{align}\label{eq:parameterization-u}
    u^{[s]}_{1:T} = \mu_{u,1:T} + x^{[s]}_{u,1:T}
\end{align}
with
\begin{align}\label{eq:parameterization-u2}
    x^{[s]}_{u,1:T} = B_u^{-1} \epsilon^{[s]}_{u,1:T} \quad\text{where}\quad  \epsilon^{[s]}_{u,1:T} \sim \mathcal N(0, I).
\end{align}
We obtain $x^{[s]}_{u,1:T}$ in $\Theta(T)$ time by solving the bidiagonal linear system $B_u x^{[s]}_{u,1:T} = \epsilon^{[s]}_{u,1:T}$.
Samples $v^{[s]}_{1:T}$ for the context embedding trajectories are obtained analogously.
Our implementation uses $S=1$, i.e., we draw only a single sample per training step.
Averaging over several samples is done implicitly since we calculate the update steps using the Adam optimizer~\citep{kingma_adam_2014}, which effectively averages over several gradient estimates in its first moment estimate.

The derivatives of $\mathcal L$ with respect to $\mu_{u,1:T}$ can be obtained using Eq.~\ref{eq:parameterization-u} and the chain rule.
We find
\begin{align}\label{eq:grad-mu}
    \frac{\partial\mathcal L}{\partial\mu_{u,1:T}} &\approx \frac{1}{S} \sum_{s=1}^S \left[ \Gamma^{[s]}_{u,1:T} - \Pi u_{1:T}^{[s]} \right].
\end{align}
Here, $\Pi\in\mathbb R^{T\times T}$ is the precision matrix of the prior $u_{1:T} \sim\mathcal N(0,\Pi^{-1})$.
It is tridiagonal and therefore the matrix-multiplication $\Pi u_{1:T}^{[s]}$ can be carried out efficiently.
The non-zero matrix elements of $\Pi$ are
\begin{align}
    \Pi_{11} &= \sigma_0^{-2} + \sigma_1^{-2} \nonumber\\
    \Pi_{TT} &= \sigma_0^{-2} + \sigma_{T-1}^{-2} \nonumber\\
    \Pi_{tt} &= \sigma_0^{-2} + \sigma_{t-1}^{-2} + \sigma_{t}^{-2}\quad \forall t\in\{2,\ldots, T-1\} \nonumber\\
    \Pi_{1,t+1} &= \Pi_{t+1,1} = -\sigma_t^{-2}. \label{eq:prior-precision}
\end{align}
The term $\Gamma_{u,1:T}^{[s]}$ on the right-hand side of Eq.~\ref{eq:grad-mu} comes from the expectation value of the log-likelihood under $q$.
It is given by
\begin{align}
    \Gamma^{[s]}_{ui,t} &= \sum_{j=1}^L \Big[\!\left(
        n_{ij,t}^+ + n_{ij,t}^-\right) \sigma\!\left(-u^{[s]\top}_{i,t} v^{[s]}_{j,t}\right) - n_{ij,t}^- \Big]  v_{j,t}^{[s]} \label{eq:Gamma}
\end{align}
where we temporarily restored the indices $i$ and $j$ for words and contexts, respectively.
In deriving Eq.~\ref{eq:Gamma}, we used the relations $\partial\log\sigma(x)/\partial x = \sigma(-x)$ and $\sigma(-x) = 1-\sigma(x)$.

The derivatives of $\mathcal L$ with respect to $\nu_{u,t}$ and $\omega_{u,t}$ are more intricate.
Using the parameterization in Eqs.~\ref{eq:parameterization-u}--\ref{eq:parameterization-u2}, the derivatives are functions of $\partial B_u^{-1}/\partial \nu_{t}$ and $\partial B_u^{-1}/\partial \omega_{t}$, respectively, where $B_u^{-1}$ is a dense (upper triangular) $T\times T$ matrix.
An efficient way to obtain these derivatives is to use the relation
\begin{align}\label{eq:grad-bmatrix}
    \frac{\partial B_u^{-1}}{\partial \nu_t} = - B_u^{-1} \frac{\partial B_u}{\partial \nu_t} B_u^{-1}
\end{align}
and similarly for $\partial B_u^{-1}/\partial \omega_{t}$.
Using this relation and Eqs.~\ref{eq:parameterization-u}--\ref{eq:parameterization-u2}, we obtain the gradient estimates
\begin{align}
    \frac{\partial\mathcal L}{\partial\nu_{u,t}} &\approx -\frac{1}{S} \sum_{s=1}^S y^{[s]}_{u,t} x^{[s]}_{u,t} - \frac{1}{\nu_{u,t}}, \label{eq:grad-nu}\\
    \frac{\partial\mathcal L}{\partial\omega_{u,t}} &\approx -\frac{1}{S} \sum_{s=1}^S y^{[s]}_{u,t} x^{[s]}_{u,t+1}. \label{eq:grad-omega}
\end{align}
The second term on the right-hand side of Eq.~\ref{eq:grad-nu} is the derivative of the entropy, Eq.~\ref{eq:def-entropy}, and
\begin{align}
    y_{u,1:T}^{[s]} &= (B_u^{\top})^{-1}\,\left[\Gamma^{[s]}_{u,1:T} -  \Pi u_{1:T}^{[s]}\right]. \label{eq:def-y}
\end{align}
The values $y_{u,1:T}^{[s]}$ can again be obtained in $\Theta(T)$ time by bringing $B_u^\top$ to the left-hand side and solving the corresponding bidiagonal linear system of equations.

\paragraph{Sampling in vocabulary space.}
In the above paragraph, we described an efficient strategy to obtain gradient estimates in only $\Theta(T)$ time.
However, the gradient estimation scales quadratic in the vocabulary size $L$ because all $L^2$ elements of the positive count matrices $n^+_t$ contribute to the gradients.
In order speed up the optimization, we pretrain the model using a minibatch of size $L'<L$ in vocabulary space as explained below.
The computational complexity of a single training step in this setup scales as $(L')^2$ rather than $L^2$.
After $N'_\text{tr}=5000$ training steps with minibatch size $L'$, we switch to the full batch size of $L$ and train the model for another $N_\text{tr}=1000$ steps.

The subsampling in vocabulary space works as follows.
In each training step, we independently draw a set $\mathcal I$ of $L'$ random distinct words and a set $\mathcal J$ of $L'$ random distinct contexts from a uniform probability over the vocabulary.
We then estimate the gradients of $\mathcal L$ with respect to only the variational parameters that correspond to words $i\in\mathcal I$ and contexts $j \in \mathcal J$.
This is possible because both the prior of our dynamic skip-gram model and the variational distribution factorize in the vocabulary space.
The likelihood of the model, however, does not factorize.
This affects only the definition of $\Gamma_{uik,t}^{[s]}$ in Eq.~\ref{eq:Gamma}.
We replace $\Gamma^{[s]}_{uik,t}$ by an estimate $\Gamma^{[s]\prime}_{uik,t}$ based on only the contexts $j\in\mathcal J$ in the current minibatch,
\begin{align}
    \Gamma^{[s]}_{ui,t} &= \frac{L}{L'} \sum_{j\in\mathcal J} \Big[\left(
        n_{ij,t}^+ + n_{ij,t}^-\right)\, \sigma\!\left(-u^{[s]\top}_{i,t} v^{[s]}_{j,t}\right) \nonumber\\[-5pt]
        &\qquad\qquad\quad - n_{ij,t}^- \Big] \, v_{j,t}^{[s]}. \label{eq:Gamma-subsample}
\end{align}
Here, the prefactor $L/L'$ restores the correct ratio between evidence and prior knowledge \citep{hoffman_stochastic_2013}.

\paragraph{Enforcing positive definiteness.}
We update the variational parameters using stochastic gradient ascent with the Adam optimizer~\citep{kingma_adam_2014}.
The parameters $\nu_{u,1:T}$ are the eigenvalues of the matrix $B_u$, which is the Cholesky decomposition of the precision matrix of $q$.
Therefore, $\nu_{u,t}$ has to be positive for all $t\in\{1,\ldots,T\}$.
We use mirror ascent~\citep{ben_ordered_2001,beck_mirror_2003} to enforce $\nu_{u,t}>0$.
Specifically, we update $\nu_t$ to a new value $\nu_t'$ defined by
\begin{align}\label{eq:mirrorascent}
    \nu_{u,t}' = \frac{1}{2} \nu_{u,t} d[\nu_{u,t}] + \sqrt{\left(\frac{1}{2} \nu_{u,t} d[\nu_{u,t}] \right)^2 + \nu_{u,t}^2}
\end{align}
where $d[\nu_{u,t}]$ is the step size obtained from the Adam optimizer.
Eq.~\ref{eq:mirrorascent} can be derived from the general mirror ascent update rule $\Phi'(\nu_{u,t}') = \Phi'(\nu_{u,t}) + d[\nu_{u,t}]$ with the mirror map $\Phi: x\mapsto -c_1\log(x) + c_2 x^2/2$, where we set the parameters to $c_1=\nu_{u,t}$ and $c_2 = 1/\nu_{u,t}$ for dimensional reasons.
The update step in Eq.~\ref{eq:mirrorascent} increases (decreases) $\nu_{u,t}$ for positive (negative) $d[\nu_{u,t}]$, while always keeping its value positive.

\paragraph{Natural basis.}
As a final remark, let us discuss an optional extension to the skip-gram smoothing algorithm that converges in less training steps.
This extension only increases the efficiency of the algorithm.
It does not change the underlying model or the choice of variational distribution.
Observe that the prior of the dynamic skip-gram model connects only neighboring time-steps with each other.
Therefore, the gradient of $\mathcal L$ with respect to $\mu_{u,t}$ depends only on the values of $\mu_{u,t-1}$ and $\mu_{u,t+1}$.
A naive implementation of gradient ascent would thus require $T-1$ update steps until a change of $\mu_{u,1}$ affects updates of $\mu_{u,T}$.

This problem can be avoided with a change of basis from $\mu_{u,1:T}$ to new parameters $\rho_{u,1:T}$,
\begin{align}\label{eq:reparameterization-mu}
    \mu_{u,1:T} = A \rho_{u,1:T}
\end{align}
with an appropriately chosen invertible matrix $A\in\mathbb R^{T\times T}$.
Derivatives of $\mathcal L$ with respect to $\rho_{u,1:T}$ are given by the chain rule, $\partial\mathcal L/\partial \rho_{u,1:T} = (\partial\mathcal L/\partial \mu_{u,1:T}) A$.
A natural (but inefficient) choice for $A$ is to stack the eigenvectors of the prior precision matrix $\Pi$, see Eq.~\ref{eq:prior-precision}, into a matrix.
The eigenvectors of $\Pi$ are the Fourier modes of the Kalman filtering prior (with a regularization due to $\sigma_0$).
Therefore, there is a component $\rho_{u,t}$ that corresponds to the zero-mode of $\Pi$, and this component learns an average word embedding over all times.
Higher modes correspond to changes of the embedding vector over time.
A single update to the zero immediately affects all elements of $\mu_{u,1:T}$, and therefore changes the word embeddings at all time steps.
Thus, information propagates quickly along the time dimension.
The downside of this choice for $A$ is that the transformation in Eq.~\ref{eq:reparameterization-mu} has complexity $\Omega(T^2)$, which makes update steps slow.

As a compromise between efficiency and a natural basis, we propose to set $A$ in Eq.~\ref{eq:reparameterization-mu} to the Cholesky decomposition of the prior covariance matrix $\Pi^{-1}\equiv A A^\top$.
Thus, $A$ is still a dense (upper triangular) matrix, and, in our experiments, updates to the last component $\rho_{u,T}$ affect all components of $\mu_{u,1:T}$ in an approximately equal amount.
Since $\Pi$ is tridiagonal, the inverse of $A$ is bidiagonal, and Eq.~\ref{eq:reparameterization-mu} can be evaluated in $\Theta(T)$ time by solving $A \mu_{u,1:T}=\rho_{u,1:T}$ for $\mu_{u,1:T}$.
This is the parameterization we used in our implementation of the skip-gram smoothing algorithm.

\bibliography{references} \bibliographystyle{icml2017}